\documentclass[runningheads]{llncs}
\usepackage[T1]{fontenc}
\usepackage{graphicx}
\usepackage{multirow}
\usepackage{booktabs}
\usepackage{amsmath}
\usepackage{colortbl}
\usepackage{pifont}
\usepackage{xcolor}
\usepackage{tabularx}
\usepackage{amssymb}
\usepackage{placeins}
\usepackage{float}
\newcommand{\cmark}{\textcolor{green!60!black}{\ding{51}}}
\newcommand{\xmark}{\textcolor{red!80!black}{\ding{55}}}
\begin{document}
\title{A Practical Guide Towards Interpreting Time-Series Deep Clinical Predictive Models: A Reproducibility Study}
%
%
\author{Yongda Fan\inst{1,2}$^*$ \and
John Wu\inst{1,2}$^*$ \and
Andrea Fitzpatrick\inst{1,2} \and
Naveen Baskaran\inst{1,2} \and
Jimeng Sun\inst{1} \and
Adam Cross\inst{3}}
\authorrunning{Y. Fan et al.}
%
\institute{University of Illinois Urbana-Champaign, Champaign, IL 61801, USA
\email{\{yongdaf2,johnwu3\}@illinois.edu} \and
PyHealth \and
University of Illinois College of Medicine, Chicago, IL 60612, USA}
\maketitle              

\begin{abstract}
Clinical decisions are high-stakes and require explicit justification, making model interpretability essential for auditing deep clinical models prior to deployment. As the ecosystem of model architectures and explainability methods expands, critical questions remain: Do architectural features like attention improve explainability? Do interpretability approaches generalize across clinical tasks? While prior benchmarking efforts exist, they often lack extensibility and reproducibility, and critically, fail to systematically examine how interpretability varies across the interplay of clinical tasks and model architectures. To address these gaps, we present a comprehensive benchmark evaluating interpretability methods across diverse clinical prediction tasks and model architectures. Our analysis reveals that: (1) attention when leveraged properly is a highly efficient approach for faithfully interpreting model predictions; (2) black-box interpreters like KernelSHAP and LIME are computationally infeasible for time-series clinical prediction tasks; and (3) several interpretability approaches are too unreliable to be trustworthy. From our findings, we discuss several guidelines on improving interpretability within clinical predictive pipelines. To support reproducibility and extensibility, we provide our implementations via PyHealth, a well-documented open-source framework: \url{https://github.com/sunlabuiuc/PyHealth}.


\end{abstract}
\section{Introduction}
One of the key barriers to deploying AI models in clinical settings is the need for explainability in deep clinical predictive models, as identified by practicing clinicians \cite{he2019practical_implementation_ai}. Beyond practical concerns such as model trustworthiness, regulations governing automated clinical systems legally require justification for each automated decision \cite{abgrall2024should_xai_govt}. This has spurred development of interpretability methods ranging from white-box approaches like mechanistic interpretability \cite{bereska2024mechanisticinterpretabilityaisafety} and gradient-based methods \cite{ancona2019gradient_based_attr} to black-box approaches like SHAP \cite{lundberg2017unifiedapproachinterpretingmodel_shap}. While several studies have explored these approaches within the clinical domain \cite{teng2022survey_clinical_deep_interp}, the best approach for interpreting deep clinical predictive models still remains unclear \cite{johannssen2025crucial_not_clear_benchmarks}. 

To better understand this problem, we devise an interpretability benchmark, and identify two practical concerns that guide our evaluation of interpretability approaches:

\textbf{Scalability across patient events and populations.} Understanding model predictions extends beyond analyzing single samples. It requires exploring diverse feature combinations and characterizing the model's prediction space across entire populations and various modalities. Interpretability approaches must therefore scale to patient populations that contain hundreds of thousands of patients and millions of clinical events. 

\textbf{Faithfulness to downstream predictions.} While explanations may not always provide immediately useful qualitative insights \cite{cina2025we_need_xai_forhealth}, they must demonstrably influence the model's predictions. Explanations that fail to affect model outputs are fundamentally untrustworthy.

With these criteria, we address two key questions in our reproducibility study:

\textbf{Do architectural changes such as attention improve explainability?} Three model families dominate clinical time-series prediction \cite{morid2023time_survey_ai4h}\cite{wang2024recent_survey}: state-based recurrent models like StageNet \cite{gao2020stagenet}, attention-based models like Transformers \cite{vaswani2017attention}, and hybrid architectures combining both approaches. Beyond improving downstream performance, attention mechanisms are often claimed to enhance model interpretability \cite{pandey2023interpretabilityattentionnetworks}, motivating numerous attention-focused interpretability methods \cite{chefer2021transformer_attn}. We directly compare attention-based and non-attention-based models to assess their impact on explanation faithfulness.

\textbf{Do interpretability approaches generalize across tasks?} An interpretability method effective for one task may fail when input and output distributions differ. We evaluate how interpretability approaches perform across diverse clinical prediction tasks, including length of stay, mortality prediction, and condition-specific predictions such as diabetic ketoacidosis.

In exploring these questions, our contributions are: (1) we demonstrate that attention, when leveraged properly, is an efficient and effective tool for interpretability across all tasks; (2) we show that black-box interpreters like KernelSHAP and LIME are computationally infeasible for time-series clinical prediction tasks; and (3) we reveal that several interpretability approaches are too unreliable to be trustworthy. From our findings, we discuss several guidelines on improving interpretability within clinical predictive pipelines. To support reproducibility and extensibility, we provide our implementations via PyHealth, a well-documented open-source framework.
\begin{table*}[t]
\centering
\caption{Comparison of interpretability reproducibility studies in healthcare AI. This benchmark advances prior work by (1) evaluating recent interpretability methods, (2) comparing across multiple tasks and models, and (3) providing an accessible, open-source implementation in PyHealth that can directly extend to workflows beyond this study.}
\label{tab:comparison}
\small
\begin{tabularx}{\textwidth}{l*{5}{>{\centering\arraybackslash}X}}
\toprule
\textbf{Study} & 
\textbf{Extensible to Other Workflows} & 
\textbf{Public Code Available} & 
\textbf{Explores Different Tasks} & 
\textbf{Cross-compares Models \& Tasks} &
\textbf{Explores New Approaches} \\
\midrule
BenchXAI \cite{metsch2025benchxai} & \xmark & \cmark & \cmark & \xmark & \xmark \\
\midrule
MIMIC-IF \cite{meng2022interpretability_mimic_if} & \xmark & \xmark & \xmark & \xmark & \xmark \\
\midrule
Zhou et al. \cite{zhou2025benchmarking_interp_pattern_discovery} & \xmark & \xmark & \xmark & \xmark & \xmark \\
\midrule
Brankovic et al. \cite{brankovic2024benchmarking_two_most_popular} & \xmark & \xmark & \cmark & \xmark & \xmark \\
\midrule
\textbf{Ours} & \textbf{\cmark} & \textbf{\cmark} & \textbf{\cmark} & \textbf{\cmark} & \textbf{\cmark} \\
\bottomrule
\end{tabularx}
\end{table*}

\section{Related Works}


\textbf{Existing interpretability benchmarks.} While previous work has benchmarked interpretability approaches on clinical tasks \cite{meng2022interpretability_mimic_if}\cite{zhou2025benchmarking_interp_pattern_discovery}\cite{brankovic2024benchmarking_two_most_popular}\cite{metsch2025benchxai}, these efforts have notable limitations. First, they inadequately explore task diversity and architectural biases, often evaluating interpretability methods on a single task or model architecture \cite{meng2022interpretability_mimic_if}\cite{zhou2025benchmarking_interp_pattern_discovery}\cite{brankovic2024benchmarking_two_most_popular}. Second, most focus primarily on older post-hoc techniques like LIME and SHAP, neglecting modern attention-based mechanisms \cite{meng2022interpretability_mimic_if}\cite{zhou2025benchmarking_interp_pattern_discovery}\cite{brankovic2024benchmarking_two_most_popular}\cite{metsch2025benchxai}. Third, many lack publicly available code, hindering reproducibility \cite{meng2022interpretability_mimic_if}\cite{zhou2025benchmarking_interp_pattern_discovery}\cite{brankovic2024benchmarking_two_most_popular}. 

In Table \ref{tab:comparison}, our framework addresses these gaps in three key ways. Unlike existing benchmarks that focus purely on measuring interpretability performance, we provide an extensible toolkit that researchers can directly integrate into custom clinical workflows through PyHealth. We systematically evaluate interpretability across multiple tasks, model architectures, and both traditional post-hoc methods and modern attention mechanisms. Given recent advances in interpretability methods, we believe such an accessible framework for systematic evaluation is urgently needed.

\section{Methodology}
There are a massive number of interpretability approaches \cite{nazir2023survey_interp_imaging} that have been developed, making extensive testing of all interpretability approaches out of scope for our work. Nonetheless, there are several key approaches that are classically used in many other interpretability evaluations. 

\textbf{Setup.} All interpretability methods evaluated in this work share a common goal: given an input $\mathbf{x} \in \mathbb{R}^d$ and model $f$, produce an attribution map $\mathbf{A} \in \mathbb{R}^d$ where each element $A_i$ represents the importance of feature $x_i$ to the prediction $f(\mathbf{x})$. This common output format enables fair comparison across diverse attribution strategies---from Shapley values to gradient flows to attention mechanisms---which differ primarily in their theoretical justification and computational approach for assigning these importance weights.

\textbf{Black-box interpreters.} Due to their model agnostic nature, black-box interpreters are a very popular first choice when attempting to interpret a clinical predictive model \cite{guidotti2018survey_black_box_interpreters}. Of particular note, LIME \cite{ribeiro2016model_lime} and SHAP \cite{lundberg2017unifiedapproachinterpretingmodel_shap} are well cited amongst the literature. Both SHAP and LIME are additive feature attribution methods that explain a prediction $f(x)$ via a linear explanation model $g(z') = \phi_0 + \sum_{i=1}^{M} \phi_i z'_i$, where $z' \in \{0,1\}^M$ indicates feature presence and each $\phi_i$ is a feature's contribution. LIME estimates the $\phi_i$ by fitting a weighted linear regression around the input with a heuristically chosen kernel $\pi_{x'}$. SHAP instead computes Shapley values from cooperative game theory:
\[
\phi_i = \sum_{S \subseteq F \setminus \{i\}} \frac{|S|!\,(|F|-|S|-1)!}{|F|!} \left[ f_{S \cup \{i\}}(x_{S \cup \{i\}}) - f_S(x_S) \right],
\]
which are the unique attributions satisfying local accuracy, missingness, and consistency. Here, we implement Kernel SHAP, which unifies the two by showing LIME recovers exact Shapley values under a specific kernel $\pi_{x'}(z') = (M-1) / \left[\binom{M}{|z'|}\,|z'|\,(M - |z'|)\right]$ with no regularization.

\textbf{Gradient-Based Counterfactual Attribution.} A growing class of attribution methods explain predictions by measuring how the output changes as inputs move from a reference state to their observed values, differing primarily in how they compute and propagate these changes. \textbf{Integrated Gradients} \cite{sundararajan2017axiomaticattributiondeepnetworks_ig} accumulates continuous gradients along a straight-line path from baseline $x^0$ to input $x$:
\[
\text{IG}_i(x) = (x_i - x_i^0) \int_0^1 \frac{\partial F(x^0 + \alpha(x - x^0))}{\partial x_i} \, d\alpha,
\]
satisfying completeness ($\sum_i \text{IG}_i = F(x) - F(x^0)$) and implementation invariance. In practice, the integral is approximated via $m$ interpolation steps, each requiring a gradient computation. \textbf{DeepLIFT} \cite{shrikumar2017learning_deeplift} achieves a similar summation-to-delta property in a single forward-backward pass by propagating discrete activation differences layer-by-layer. This is far cheaper, but because the chain rule does not hold for discrete gradients, DeepLIFT can yield different attributions for functionally equivalent networks, violating implementation invariance \cite{sundararajan2017axiomaticattributiondeepnetworks_ig}.

Both methods can underestimate component importance in transformers due to \emph{self-repair}, where downstream components compensate for perturbations. \cite{edin2025gimimprovedinterpretabilitylarge_llm} observe that this is especially prevalent in LLM attention mechanisms --- softmax redistribution masks the true influence of attention scores --- and propose \textbf{GIM}, which modifies gradient flow through softmax, layer normalization, and multiplicative interactions to account for these effects, yielding more faithful attributions.

\textbf{Attention-based Attribution.} Finally, from the transformer architecture came a variety of works that claim to improve the interpretability of their models through the attention mechanism\cite{pandey2023interpretabilityattentionnetworks} \cite{chefer2021transformer_attn} \cite{choi2016retain}\cite{mullenbach2018explainablepredictionmedicalcodes_caml}. Of particular note, \textbf{Chefer} \cite{chefer2021transformer_attn} shows that by simply aggregating and weighing the attention maps with its gradients, they can dramatically improve the faithfulness and explainability of transformer models compared to simply using only their attention scores.

Ultimately, each approach has shown valid empirical evidence of their utility \cite{xu2025interpretability_survey}, making their exploration in a fair and reproducible manner a key priority. 
\section{Results}
\begin{table*}[h]
\newcommand{\colw}[1]{\makebox[3.5em]{#1}}
\centering
\caption{Interpretability Performance Matrix. Each cell shows the number of model-task pairs where the row method outperforms the column method, scored by $\text{Comprehensiveness} \times (1 - \text{Sufficiency})$. \textbf{Darker orange indicates higher win rate when reading horizontally, and higher lose rates when reading vertically.} }
\label{tab:interpret_matrix}
\begin{tabular}{l|l|ccccccc}
\toprule
\multicolumn{2}{c}{} & \multicolumn{7}{c}{\textbf{Lose}} \\
\cmidrule{3-9}
\multicolumn{2}{c}{} & \colw{Chefer} & \colw{DeepLift} & \colw{GIM} & \colw{IG} & \colw{LIME} & \colw{SHAP} & \colw{Baseline} \\
\cmidrule{3-9}
\multirow{7}{*}{\textbf{Win}}
& Chefer   & \cellcolor{gray!20}-   & \cellcolor{orange!70}5/6 & \cellcolor{orange!100}6/6 & \cellcolor{orange!50}4/6 & \cellcolor{orange!70}5/6 & \cellcolor{orange!70}5/6 & \cellcolor{orange!100}6/6 \\
& DeepLift & \cellcolor{orange!10}1/6 & \cellcolor{gray!20}-   & \cellcolor{orange!35}4/9 & \cellcolor{orange!10}1/9 & \cellcolor{orange!45}5/9 & \cellcolor{orange!25}3/9 & \cellcolor{orange!55}6/9 \\
& GIM      & \cellcolor{orange!0}0/6 & \cellcolor{orange!45}5/9 & \cellcolor{gray!20}-   & \cellcolor{orange!10}1/9 & \cellcolor{orange!45}5/9 & \cellcolor{orange!25}3/9 & \cellcolor{orange!65}7/9 \\
& IG       & \cellcolor{orange!25}2/6 & \cellcolor{orange!80}8/9 & \cellcolor{orange!80}8/9 & \cellcolor{gray!20}-   & \cellcolor{orange!100}9/9 & \cellcolor{orange!65}7/9 & \cellcolor{orange!100}9/9 \\
& LIME     & \cellcolor{orange!10}1/6 & \cellcolor{orange!35}4/9 & \cellcolor{orange!35}4/9 & \cellcolor{orange!0}0/9 & \cellcolor{gray!20}-   & \cellcolor{orange!15}2/9 & \cellcolor{orange!55}6/9 \\
& SHAP     & \cellcolor{orange!10}1/6 & \cellcolor{orange!55}6/9 & \cellcolor{orange!55}6/9 & \cellcolor{orange!15}2/9 & \cellcolor{orange!65}7/9 & \cellcolor{gray!20}-   & \cellcolor{orange!100}9/9 \\
& Baseline & \cellcolor{orange!0}0/6 & \cellcolor{orange!25}3/9 & \cellcolor{orange!15}2/9 & \cellcolor{orange!0}0/9 & \cellcolor{orange!25}3/9 & \cellcolor{orange!0}0/9 & \cellcolor{gray!20}-   \\
\bottomrule
\end{tabular}
\end{table*}

\textbf{Dataset.} We evaluate interpretability approaches on three MIMIC-IV clinical tasks \cite{johnson2023mimic}: diabetic ketoacidosis (DKA) prediction, mortality prediction, and length-of-stay prediction. Following the StageNet implementation \cite{gao2020stagenet}, we prepare patient-level data using ICD codes and lab events as features, including both time intervals and clinical measurements. The datasets contain 137,778 patients (mortality), 220,853 samples (length-of-stay), and 179,945 samples (DKA). Due to the computational complexity of SHAP and LIME, we interpret approximately 1,000 randomly selected samples for fair comparison across all methods and tasks. Dataset details are available in our codebase.

\textbf{Models.} We train three models for each task: StageNet \cite{gao2020stagenet}, Transformer \cite{vaswani2017attention}, and StageAttn—a modified StageNet with an additional multi-head attention layer to examine attention's effect on interpretability. All models achieve comparable performance on mortality and DKA prediction. However, on length-of-stay prediction, the Transformer achieves only half the accuracy of StageNet and StageAttn. Performance metrics for all model-task pairs are in Table~\ref{tab:model_performance_all}.

\textbf{Baselines.} We evaluate six interpretability methods---Chefer, DeepLIFT, GIM (temperature 2.0), Integrated Gradients (50 steps), LIME (200 samples), and Kernel SHAP---across all model-task pairs. A random baseline assesses whether attribution methods provide meaningful signal by randomly highlighting features.

\begin{figure}[h!]
    \centering
    \includegraphics[width=1.0\textwidth]{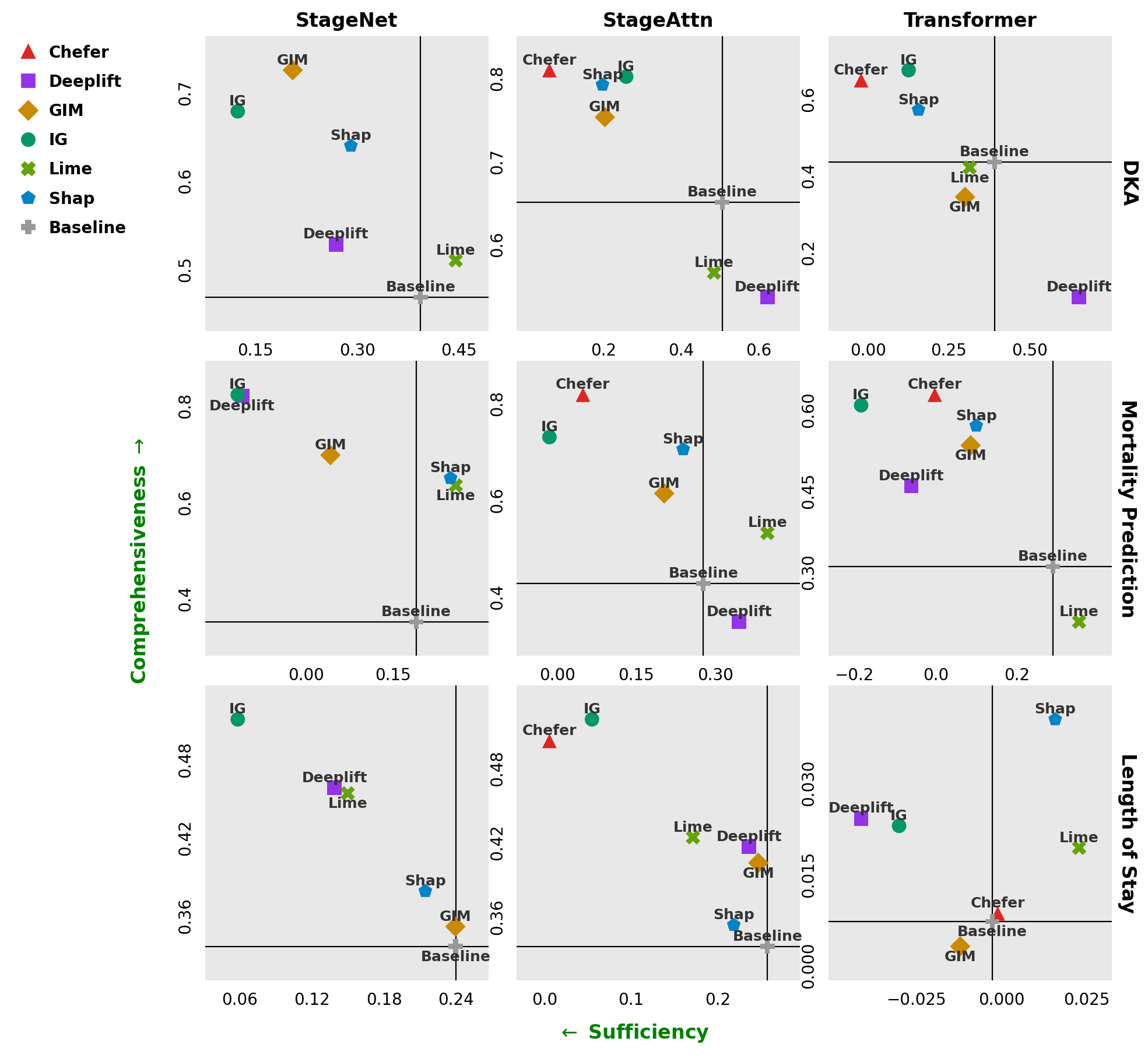}
    \caption{\textbf{Faithfulness Benchmark Results.} For proper interpretation, (green arrows) higher comprehensiveness and lower sufficiency imply more faithful explanations generated. We observe that Integrated Gradients is consistently a top-performer in terms of interpretation and Chefer emerges as a competitive method for attention-based models, while many of the other approaches move a tremendous degree depending on the task and model. }
    \label{fig:evaluation}
\end{figure}

\begin{figure}[h!]
    \centering
    \includegraphics[width=0.8\textwidth]{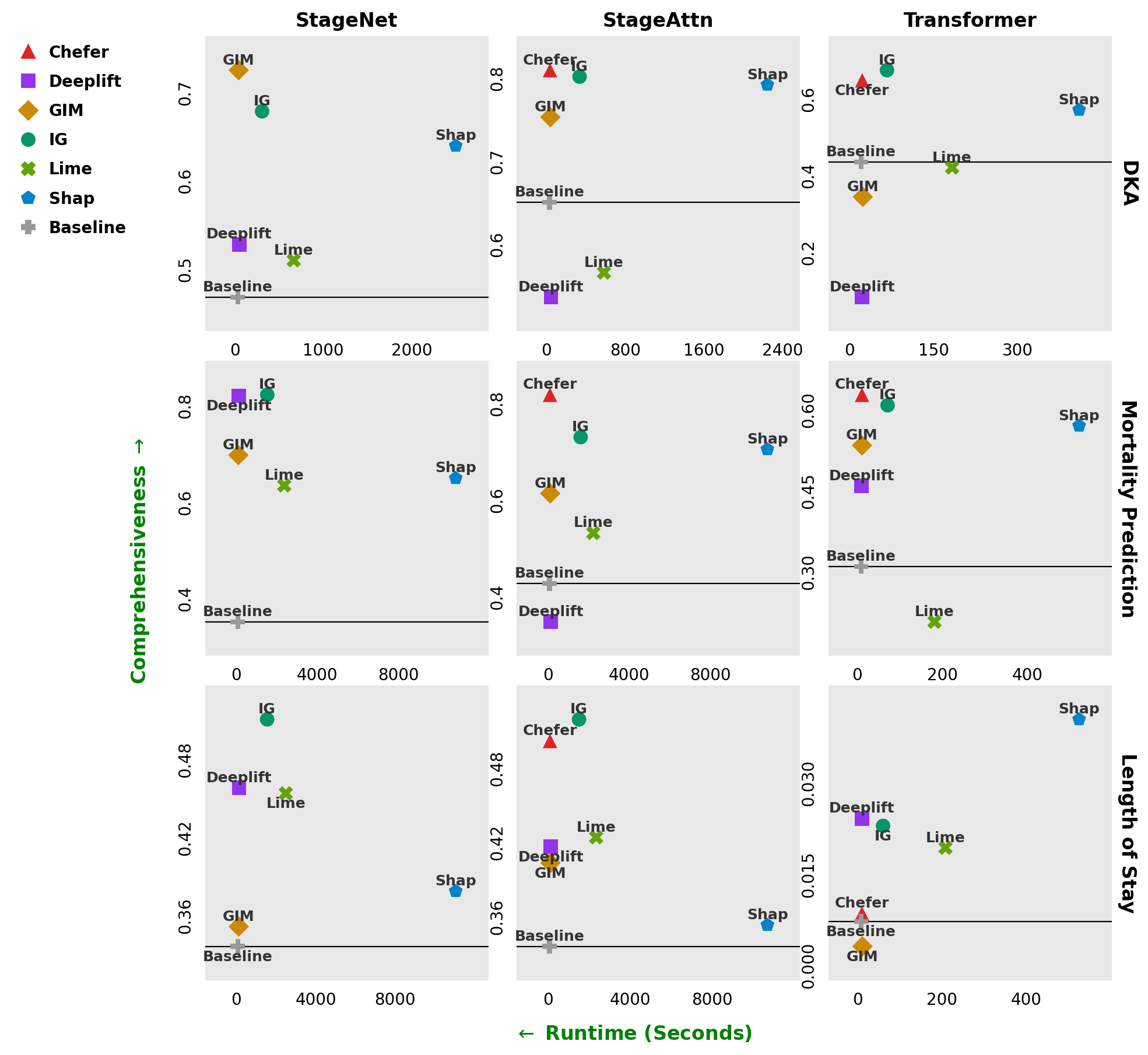}
    \caption{\textbf{Runtime Comparison.} For proper interpretation, (green arrows) higher comprehensiveness and lower runtime imply more desirable trade-off. We observe that Shap tends to run significantly longer with minimal advantages compare to others, while Integrated Gradients and Chefer can typically interpret the models better with much shorter runtime.}
    \label{fig:runtime}
\end{figure}

\textbf{Metrics.} We evaluate faithfulness using sufficiency and comprehensiveness \cite{chan2022comparativestudyfaithfulnessmetrics}, two widely-used metrics in the interpretability community. These metrics pose complementary questions: sufficiency measures the predicted softmax probability drop when removing features deemed irrelevant by an interpretability method, while comprehensiveness measures the drop when removing relevant features. Faithfulness across all tasks and models are shown in Figure~\ref{fig:evaluation} and runtime in Figure~\ref{fig:runtime}. For an overall comparison, Table~\ref{tab:interpret_matrix} reports head-to-head win rates across all model-task pairs, scored by $\text{Comprehensiveness} \times (1 - \text{Sufficiency})$.

\textbf{Top performers.} Table~\ref{tab:interpret_matrix} compares each interpretability method head-to-head across all model-task pairs. \textbf{Integrated Gradients} and \textbf{Chefer} emerge as the most faithful methods overall. Integrated Gradients consistently outperforms most approaches, and when it does not rank first, the margin to the leader is narrow. Chefer performs even more faithfully, outperforming Integrated Gradients in 4 of 6 pairs and all other methods in at least 5 of 6 pairs. Its only exception is the Transformer on length-of-stay prediction, likely due to that model's suboptimal training as shown in Table~\ref{tab:model_performance_all}. Among black-box methods, \textbf{SHAP} is the most reliable, outperforming other methods in over half of cases and consistently surpassing the random baseline, though it falls behind the gradient-based approaches.

\textbf{Unreliable methods for clinical time-series.} Three methods prove unreliable in practice. \textbf{DeepLIFT} loses to the random baseline in 3 of 9 model-task pairs, all involving attention-based models. This volatility with attention mechanisms, observed by \cite{edin2025gimimprovedinterpretabilitylarge_llm}, likely stems from its layer-by-layer design \cite{shrikumar2017learning_deeplift} conflicting with attention's self-repair mechanism \cite{edin2025gimimprovedinterpretabilitylarge_llm}. \textbf{GIM}, optimized for attention in language models \cite{edin2025gimimprovedinterpretabilitylarge_llm}, loses to the random baseline in 2 of 9 pairs and shows no advantage over other methods. It ranks first only on a non-attention model, where it practically reduces to GradientXInput \cite{sundararajan2017axiomaticattributiondeepnetworks_ig}, suggesting that substantially larger Transformers may be needed to realize its benefits \cite{edin2025gimimprovedinterpretabilitylarge_llm}. \textbf{LIME} performs worst overall, losing to the random baseline in 3 of 9 pairs and consistently trailing other methods.

\textbf{Black-box methods are computationally infeasible at scale.} Interpreting all 137,778 mortality prediction samples would require an estimated 300 hours for SHAP and 64 hours for LIME (Figure~\ref{fig:runtime}). While sampling fewer points could reduce runtime, this trades off faithfulness—an unacceptable compromise given that SHAP and LIME already underperform gradient-based methods. We recommend against black-box interpreters for deep clinical predictive models.

\textbf{Integrated Gradients: faithful but costly for non-attention models.} While 36\% faster than LIME, Integrated Gradients remains computationally expensive for large patient populations. However, its superior faithfulness (Figure~\ref{fig:evaluation}) justifies this cost for non-attention models.

\textbf{Gradient-weighed attention: the most efficient and faithful approach.} The Chefer method \cite{chefer2021transformer_attn}, which interprets aggregated gradient-weighted attention maps, consistently produces the most faithful attributions across models and tasks. It is also remarkably efficient—approximately 15 times faster than Integrated Gradients while achieving comparable faithfulness. Notably, adding attention layers to recurrent models does not harm predictive performance (Table~\ref{tab:model_performance_all}). While the full Transformer underperforms on clinical time-series tasks, StageAttn—a hybrid combining StageNet with attention—matches the original StageNet's performance. This suggests that incorporating attention layers may be a practical pathway to improving model interpretability without sacrificing accuracy.

\begin{table*}[h!]
\centering
\caption{Model Performance Across All Tasks on MIMIC-IV Dataset. Bold values indicate best performance within each task. Acc. = Accuracy, F1-W = F1-Weighted, F1-Ma = F1-Macro, F1-Mi = F1-Micro.}
\label{tab:model_performance_all}
\resizebox{\textwidth}{!}{%
\begin{tabular}{l|cccc|cccc|cccc}
\toprule
& \multicolumn{4}{c|}{\textbf{Mortality Prediction}} & \multicolumn{4}{c|}{\textbf{DKA Prediction}} & \multicolumn{4}{c}{\textbf{Length of Stay}} \\
\cmidrule(lr){2-5} \cmidrule(lr){6-9} \cmidrule(lr){10-13}
\textbf{Model} & \textbf{PR-AUC} & \textbf{ROC-AUC} & \textbf{Acc.} & \textbf{F1} & \textbf{PR-AUC} & \textbf{ROC-AUC} & \textbf{Acc.} & \textbf{F1} & \textbf{Acc.} & \textbf{F1-W} & \textbf{F1-Ma} & \textbf{F1-Mi} \\
\midrule
StageNet    & 0.6870 & \textbf{0.9576} & \textbf{0.9618} & \textbf{0.6188} & 0.0789 & \textbf{0.8604} & 0.9951 & \textbf{0.1682} & \textbf{0.5862} & \textbf{0.5809} & 0.5694 & \textbf{0.5862} \\
StageAttn   & \textbf{0.6955} & 0.9441 & 0.9602 & 0.6102 & 0.0783 & 0.8448 & 0.9963 & 0.0571 & 0.5819 & 0.5791 & \textbf{0.5741} & 0.5819 \\
Transformer & 0.6134 & 0.9488 & 0.9504 & 0.5267 & \textbf{0.1075} & 0.8400 & \textbf{0.9968} & 0.0938 & 0.2434 & 0.1800 & 0.1511 & 0.2434 \\
\bottomrule
\end{tabular}%
}
\end{table*}

\section{Future Work}
Our reproducibility study identifies two key directions for future work.

\textbf{Faithfulness across modalities.} Patient profiles are inherently multimodal, containing time-series features, imaging, and clinical notes. While we evaluate interpretability methods for clinical time-series tasks, our findings may not generalize to other modalities—for instance, GIM outperforms many baselines in language modeling \cite{edin2025gimimprovedinterpretabilitylarge_llm}. Cross-comparing interpretability approaches across models, tasks, and modalities will help identify domain-specific biases and limitations.

\textbf{Developing improved interpretability approaches.} Many interpretability methods are complementary rather than mutually exclusive. Combining these approaches may yield insights for building both more interpretable models and better interpretation methods. By releasing our benchmark in PyHealth, an extensible open-source framework, we enable others to apply and extend these techniques beyond small-scale reproducibility studies for their own tasks. 
\FloatBarrier

\begin{credits}
\subsubsection{\ackname} This study was funded by Jump ARCHES endowment
awarded by the Healthcare Engineering Systems Center at the University of Illinois
Urbana-Champaign (UIUC) and made possible through the PyHealth Research Initiative.

\subsubsection{\discintname}
There are no conflicts of interests here.

\end{credits}
\bibliographystyle{splncs04}
\bibliography{ref}




\end{document}